\documentclass[journal,twoside,web]{ieeecolor}
\usepackage{generic}
\usepackage{cite}
\usepackage{amsmath,amssymb,amsfonts}
\usepackage{algorithmic}
\usepackage{graphicx}
\usepackage{textcomp}
\usepackage[outdir=./]{epstopdf}

\graphicspath{ {images/} }
\usepackage{multirow}
\usepackage{float}

\makeatletter  
\newif\if@restonecol  
\makeatother

\usepackage[linesnumbered,ruled,vlined]{algorithm2e}

\begin{document}
\title{Deep Learning-based Biological Anatomical Landmark Detection in Colonoscopy Videos}

\author{Kaiwei Che$^1$, Chengwei Ye$^2$, Yibing Yao$^3$, Nachuan Ma$^1$, Ruo Zhang$^1$, \\ Jiankun Wang$^1$, and~Max~Q.-H.~Meng$^{124}$,~\IEEEmembership{Fellow,~IEEE}

\thanks{This work was supported by the National Key Research
and Development Program of China under Grant 2019YFB1312400. \{Kaiwei Che, Chengwei Ye, and Yibing Yao contributed equally to this work. Corresponding author:{\emph{Jiankun Wang, Max Q.-H. Meng}}\}}

\thanks{$^1$ Kaiwei Che,  Nachuan Ma, Ruo Zhang, and Jiankun Wang are with the Department of Electronic and Electrical Engineering, Southern University of Science and Technology in Shenzhen, China, {\tt\small 12032207@mail.sustech.edu.cn}, {\tt\small manachuan@163.com}, {\tt\small ruozhang0608@gmail.com}, {\tt\small wangjk@sustech.edu.cn},.}

\thanks{$^2$ Chengwei Ye is with the Department of Electronic Engineering, The Chinese University of Hong Kong, Hong Kong, {\tt\small cwye@link.cuhk.edu.hk}.}

\thanks{$^3$ Yibing Yao is with the Department of Oncology, Air Force Medical Center, PLA. No. 30 Fucheng Road, Haidian District, Beijing 100142, China, {\tt\small yaoyb777@163.com}.}

\thanks{$^{124}$ Max Q.-H. Meng is with the Department of Electronic and Electrical Engineering, Southern University of Science and Technology in Shenzhen, China, on leave from the Department of Electronic Engineering, The Chinese University of Hong Kong, Hong Kong, and also with the Shenzhen Research Institute of the Chinese University of Hong Kong in Shenzhen, China, {\tt\small max.meng@ieee.org}.}
}

\maketitle

\begin{abstract}
Colonoscopy is a standard imaging tool for visualizing the entire gastrointestinal (GI) tract of patients to capture lesion areas. However, it takes the clinicians excessive time to review a large number of images extracted from colonoscopy videos.
Thus, automatic detection of biological anatomical landmarks within the colon is highly demanded, which can help reduce the burden of clinicians by providing guidance information for the locations of lesion areas.
In this article, we propose a novel deep learning-based approach to detect biological anatomical landmarks in colonoscopy videos.
First, raw colonoscopy video sequences are pre-processed to reject interference frames. 
Second, a ResNet-101 based network is used to detect three biological anatomical landmarks separately to obtain the intermediate detection results.
Third, to achieve more reliable localization of the landmark periods within the whole video period, we propose to post-process the intermediate detection results by identifying the incorrectly predicted frames based on their temporal distribution and reassigning them back to the correct class. Finally, the average detection accuracy reaches 99.75\%. Meanwhile, the average IoU of 0.91 shows a high degree of similarity between our predicted landmark periods and ground truth. The experimental results demonstrate that our proposed model is capable of accurately detecting and localizing biological anatomical landmarks from colonoscopy videos.

\end{abstract}


\section{Introduction}

\begin{figure}[htbp]
\centerline{\includegraphics[width=21pc]{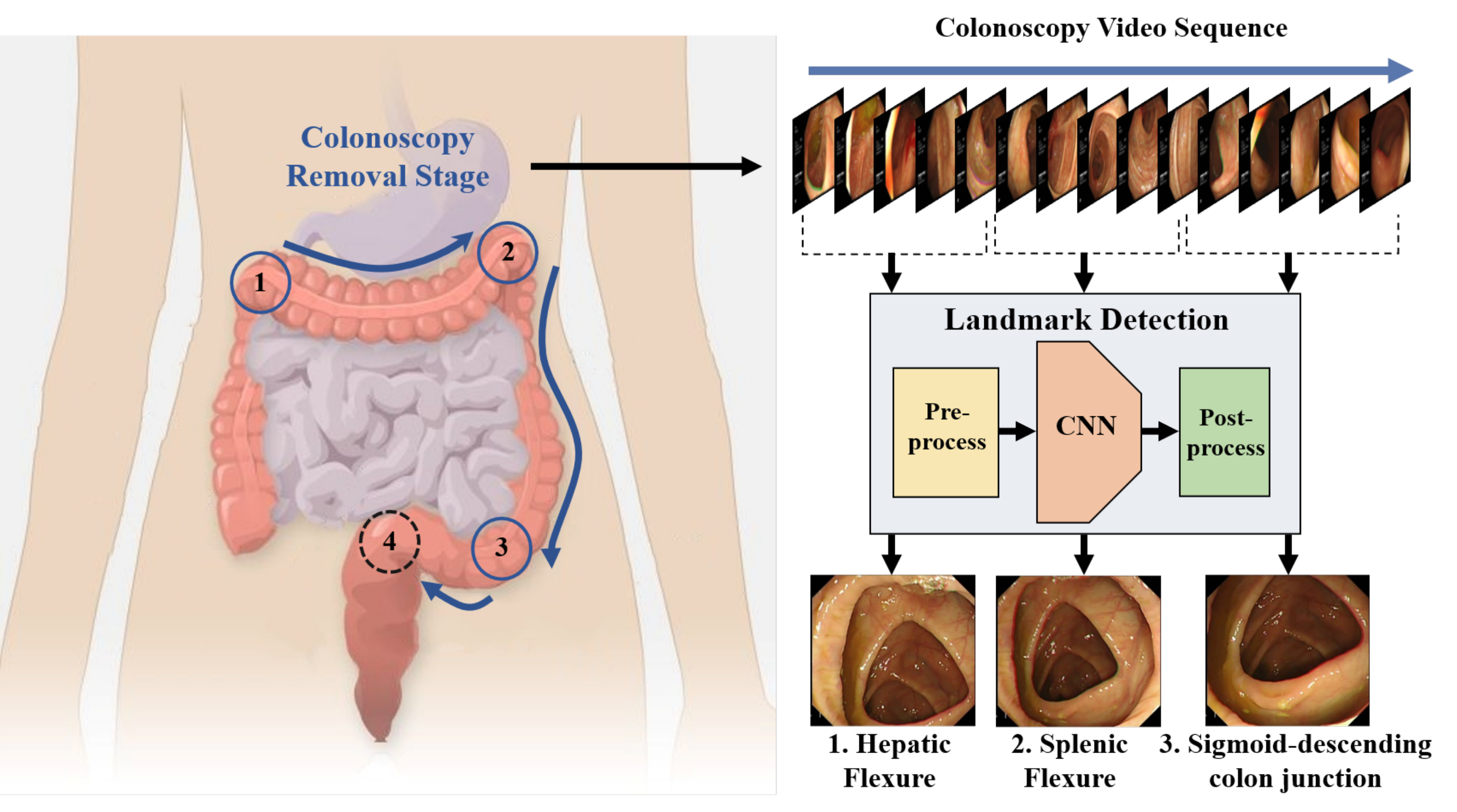}}
\caption{The four biological anatomical landmarks in colonoscopy videos are hepatic flexure, splenic flexure, sigmoid-descending colon junction, and rectosigmoid junction. The rectosigmoid junction is not among the detection objectives since it is identical to the end of the video. The colonoscopy video sequences are passed into the threefold system consisting of a pre-processing module, a detection network, and a post-processing module. The outputs of the system include prediction result for each frame and locations of the three landmark periods within the video sequence.}
\label{fig-four_landmarks}
\end{figure}

\IEEEPARstart{C}{olorectal} diseases, especially colorectal cancer (CRC), bring significant threat to public health worldwide.
According to the statistical data from the American Cancer Society, the incidence rate of CRC ranged from 30 (per 100,000 persons) in Asian to 89 in Alaska Natives from 2012 through 2016 \cite{siegel2020colorectal}.
Fortunately, CRC can be prevented if adenomatous polyps within the colon can be detected and removed in a very early stage as polyps are essential precursors to CRC \cite{siegel2017colorectal}.
Colonoscopy is one of the standard screening methods to detect polyps, with which the clinician can examine the entire rectum and colon of patients.
Clinicians make diagnostic decisions by checking the collected colonoscopy images frame by frame.
However, manual inspection of colonoscopy images is laborious and time-consuming, as images capturing the precancerous polyps occupy only a limited percentage of the complete colonoscopy images.
Moreover, clinicians cannot obtain the location of polyps within the colon directly through the collected colonoscopy videos.
Hence, automatically detecting polyps and locating them within the colon is highly demanded to ease the burden of clinicians and improve diagnostic accuracy.
In the past few years, researchers have devoted great efforts to developing automatic polyps detection methods 
\cite{zhao2011polyp,li2012automatic,jia2020automatic,yuan2017deep}.
However, though previous methods have managed to detect polyps in a robust and efficient way, there is no work to localize polyps within the colon, which is crucial in clinical practice but challenging for clinicians to do manually.
Therefore, in this article, we propose a novel deep learning-based algorithm to detect three biological anatomical landmarks in colonoscopy videos, providing a research basis for calculating the relative distances between the lesion areas (such as polyps and bleeding regions) and the landmarks.
The proposed algorithm will help reduce human error and accelerate the diagnosis process significantly.

As shown in Fig. \ref{fig-four_landmarks}, the three biological anatomical landmarks to be detected include hepatic flexure, splenic flexure, and sigmoid-descending colon junction. The fourth biological anatomical landmark is rectosigmoid junction, which requires no detection since it is the end of the colonoscopy video.

Our main contributions can be summarized in the following two aspects:
\begin{enumerate}
    \item We collect a colonoscopy video dataset and finely label the time periods of four biological anatomical landmarks for each video.
    \item We propose a novel three-fold biological anatomical landmark detection system, consisting of a pre-processing module, a deep learning-based detection network, and a post-processing module.

\end{enumerate}

The remainder of this article is organized as follows. The previous work on colon imaging data and existing biological anatomical landmark detection methods are reviewed in Section \ref{related work}. We introduce the colonoscopy video dataset in Section \ref{dataset}. Section \ref{method} outlines the proposed biological anatomical landmark detection algorithm, while the experimental results are presented and discussed in Section \ref{experimental results}. Finally, we draw some conclusions and discuss the future work in Section \ref{conclusion}.

\section{Related Work}
\label{related work}
\subsection{Previous Work on Colon Imaging Data}

Most of the previous work related to colon imaging data focuses on the detection of polyps and bleeding regions.
In previous studies, researchers tend to design hand-engineered methods based on conventional machine learning techniques and exploit low-level features such as color, texture, and shape information to detect polyps.
For instance, Zhao \textit{et al.} \cite{zhao2011polyp} converted the screening images from RGB color space to HSI/HSV color space for good property of color invariance. 
Then color histogram technique could be used to describe polyp images.
In \cite{li2012automatic}, a novel texture feature combining wavelet transform and uniform local binary pattern was utilized to describe colon imaging data. Based on that, a support vector machine (SVM) classifier was trained to automatically detect polyps.
In addition, Mamonov \textit{et al.} \cite{mamonov2014automated} assumed that polyps are always round in shape and applied this geometrical feature to image analysis for round polyps detection. 
Unfortunately, these hand-engineered methods cannot achieve satisfactory performance for their poor characterization capability. 

Recently, various deep learning methods have been applied to polyps detection with the revolution of computational technologies \cite{jia2020automatic,yuan2017deep,yuan2019densely,jia2019wireless}. 
Hierarchical feature learning and discrimination capabilities of neural networks can help significantly improve polyps detection accuracy. For instance, Jia \textit{et al.} \cite{jia2020automatic} proposed a novel two-stage architecture called PLP-Net for automatic pixel-accurate polyps detection in colonoscopy images based on deep convolutional neural network (CNN).
However, previous studies on colon imaging data did not involve localizing polyps within the colon, and further research work is demanded.
To fill this gap, we propose a novel deep learning-based algorithm to detect biological anatomical landmarks in colonoscopy videos, which provides a research foundation for localizing polyps within colon.
\subsection{Existing Biological Anatomical Landmark Detection Methods}
According to the literature\cite{zhou2019handbook}\cite{zhou2020review}, biological anatomical landmark detection plays an essential role in various medical image analysis assignments, which can help achieve registration \cite{lange20093d} and segmentation\cite{ibragimov2017segmentation} tasks of medical images.

Traditional landmark detection methods usually utilize classical machine learning algorithms and design specific image filters to extract invariant features \cite{chiras1997percutaneous,liu2010search,lindner2014robust,ebner2014towards,oktay2016stratified}. For instance, Liu \textit{et al.} \cite{liu2010search} leveraged the theory of submodular functions to search multiple human body landmarks including bone, organs, and vessels in 3D Computed Tomography (CT) images.
Lindner \textit{et al.} \cite{lindner2014robust} proposed a novel landmark detection algorithm based on the supervised random forest regression-voting method for facial landmarks detection and the annotation of the joints of the hands in radiographs.
The stratified decision forests method was also utilized to detect anatomical landmarks in cardiac images \cite{oktay2016stratified}.

Recently, researchers have proposed a large quantity of anatomical landmark detection algorithms \cite{wester2020detecting,song2020automatic,leroy2020communicative,lian2020multi,zhu2021you}  based on deep learning and reinforcement learning methods, and these algorithms showed more robust and accurate performance.
Wester \textit{et al.} \cite{wester2020detecting} used a patch-based CNN to detect anatomical landmarks in 3D cardiovascular images, providing automatic registration between ultrasound and CT images of the same patient. 
Song \textit{et al.} \cite{song2020automatic} proposed a two-step method to detect cephalometric landmarks automatically on skeletal X-ray images, utilizing pre-trained networks with a backbone of ResNet-50 \cite{he2016deep}.
Moreover, a novel communicative reinforcement learning agents system was presented for landmark detection in brain images and was evaluated on two datasets from adult magnetic resonance imaging (MRI) and fetal ultrasound scans \cite{leroy2020communicative}.

\begin{figure}[htbp]
\centerline{\includegraphics[width=16pc]{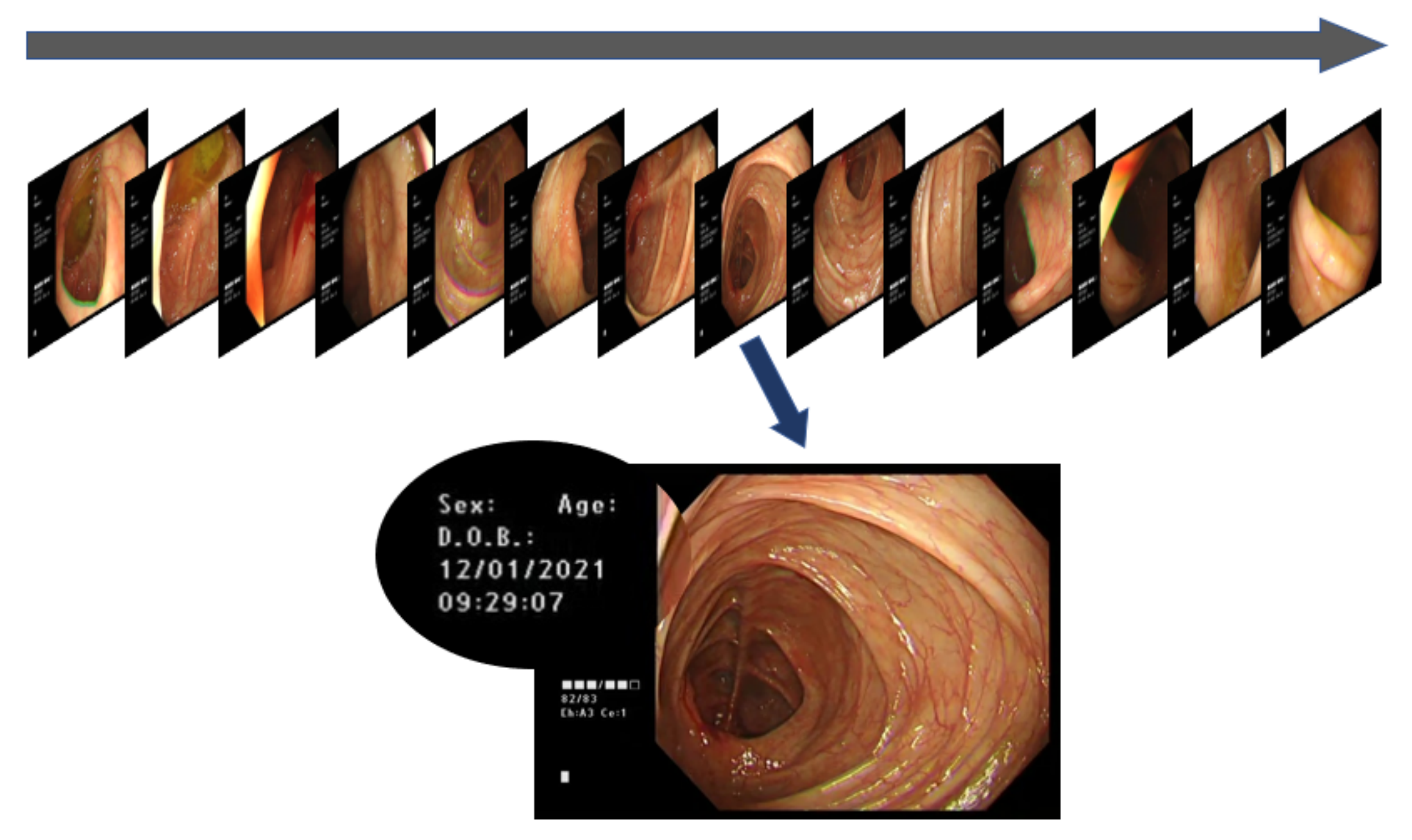}}
\caption{Each colonoscopy video is around 10-20 minutes, with clinical information and current timestamp displayed on the interface.}
\label{fig-raw_image}
\end{figure}

\begin{figure*}[htbp]
\centerline{\includegraphics[width=42pc]{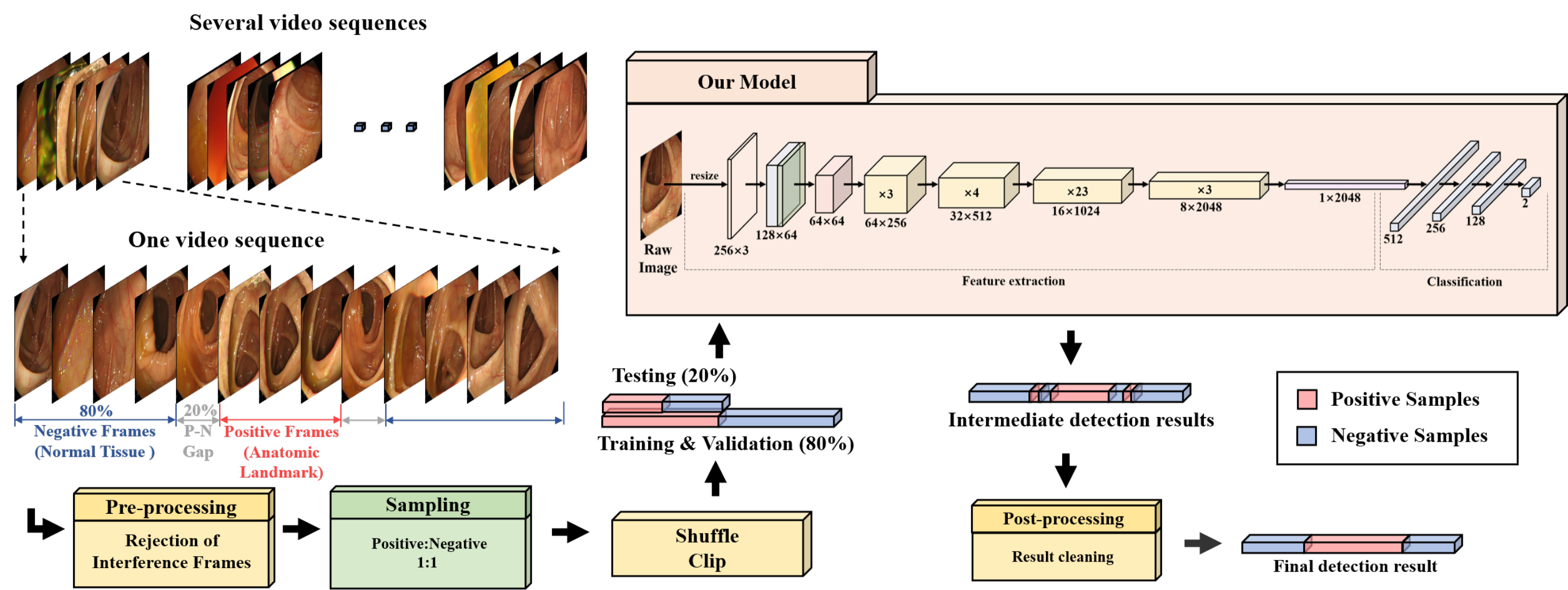}}
\caption{Workflow of the proposed landmark detection system. First, the colonoscopy video sequences are passed into the system and sampled into positive frames and negative frames separated by the P-N gap with adaptive sampling frequencies. Next, the pre-processing module is applied to reject interference frames in sampled data. The pre-processed data are then shuffled, clipped, and divided into training, validation, and test sets. The training and validation sets are applied to train and validate the detection model based on ResNet-101. After passing the test data into the trained model, the model outputs intermediate detection results indicating whether each frame should be classified as positive or negative. Finally, the post-processing module is applied to identify the incorrectly predicted frames and reassign them back to the correct class. The system outputs contain final detection results and locations of the three landmark periods within the video sequence.}
\label{fig-flow}
\end{figure*}

However, to the best of our knowledge, no research studies have worked on biological anatomical landmark detection in colonoscopy videos.
Hence, in this article, we devote to applying deep learning methods to tackle this problem.

\section{Dateset}
\label{dataset}

The dataset for biological anatomical landmark detection consists of 49 colonoscopy videos of different patients. The duration of each video is approximately 10-20 minutes, including an insertion stage and a removal stage. The videos have a resolution of $560\times720$ and a frame rate of 50fps. As shown in Fig. \ref{fig-raw_image}, clinical information including the serial number of the patient, testing date, and current timestamp are also displayed in the videos.

During both the insertion stage and the removal stage of each colonoscopy video sequence, the timestamp and scope length when passing each biological anatomical landmark are recorded by clinicians. However, due to the resistance and disturbance, while inserting the scope, videos captured from the insertion stage contain a large number of interference frames such as turbid frames and camera shaking frames. Therefore, in view of the concern for data quality during the insertion stage, ground truth labels are generated from the timestamp and scope length data during the removal stage.

As shown in Fig. \ref{fig-four_landmarks}, for each video sequence, there are three time periods during which the anatomical landmarks are detected, each with a duration of 10 to 25 seconds. The three landmark periods are represented as positive periods in ground truth labels. The labels were manually annotated and verified by expert clinicians based on the bending features and biological characteristics of the landmark regions. Since the differences between landmark regions and normal regions are subtle and hardly perceptible, identifying the landmark periods is a challenging task involving significant difficulty, which can only be accomplished by clinicians with sufficient experience.


In terms of sampling images from the videos, the sampling period for each landmark begins at the timestamp of the previous landmark (or the timestamp when the removal stage starts) and ends at the timestamp of the next landmark (or the timestamp when the removal stage ends). As shown in Fig. \ref{fig-flow}, we introduce a positive-and-negative gap (P-N gap) between the landmark period and normal tissue period to guarantee that the interclass difference is sufficiently large. Given the P-N gap, the labeled landmark period is taken as the positive period while 80\% of the two remaining parts of the sampling period are taken as the negative periods. To tackle the sampling imbalance problem, we apply adaptive sampling frequencies for landmark periods and normal tissue periods. Details are presented in Section \ref{experimental_data_and_setup}

\section{Method}
\label{method}

In a study approved by the local medical ethics committee, colonoscopy video data were obtained from 49 patients. Each patient signed an informed consent form after understanding the study. 
In this section, we first introduce a pre-processing module to reject interference frames in the collected data. Then we propose a novel network structure based on ResNet-101 to perform landmark detection. Finally, we illustrate the post-processing module for cleaning the intermediate detection results, which enables localization of landmark periods within the video period and improves final detection performance. The entire process of our proposed method is shown in Fig. \ref{fig-flow}.

\subsection{Pre-process: Rejection of Interference Frames}
 Colonoscopy video frames consist of biological anatomical landmarks frames, normal tissue frames, and interference frames caused by feces, bubbles, and camera shaking. The crucial issue of the biological anatomical landmark detection task is regarding distinguishing landmarks from normal tissue. However, during the sampling process, interference frames are likely to appear in both normal tissue sampling periods and anatomical landmark sampling periods, which will cause the detection model to take wrong samples as learning inputs and thereby reduce the detection accuracy. To facilitate the accurate detection of biological anatomical landmarks in the colon, it is necessary to pre-process all the video frames to reject those interference frames. 

\begin{figure}[htbp]
\centerline{\includegraphics[width=16pc]{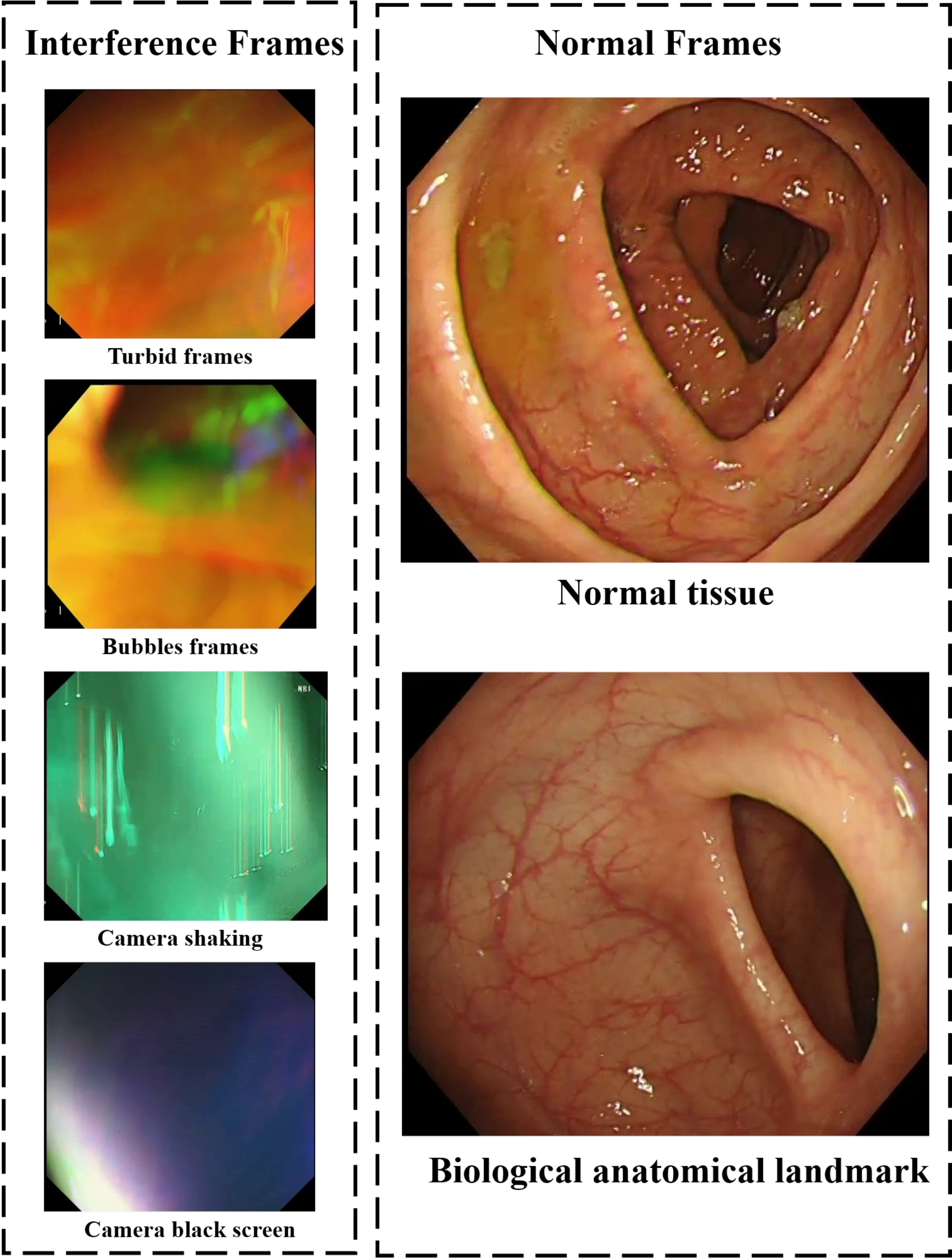}}
\caption{Interference frames and normal frames. Interference frames contain turbid frames, bubbles frames, and blurry frames resulted from camera shaking and black screen. Normal frames contain biological anatomical landmark frames and normal tissue frames.}
\label{fig-method-original_image}
\end{figure}

As shown in Fig. \ref{fig-method-original_image}, normal frames include biological anatomical landmarks and normal tissue, while interference frames include turbid frames, bubbles frames, and blurry frames caused by camera shaking and black screen. It can be observed from Fig. \ref{fig-method-original_image} that normal frames and interference frames exhibit clear distinctions, especially in terms of texture and color features. The textures of normal frames are finer and more complex than the textures of interference frames. Further, the color of normal frames involves frequent changes, whereas the interference frames are relatively monochromatic. 

Based on the observation of the distinctions between normal and interference frames, the rejection of interference frames is achieved by applying an image processing tool named Canny edge detector \cite{DING2001721}. The Canny edge detector has been widely applied in the field of computer vision to locate sudden changes in intensity and find object boundaries in images. In the direction of the maximum intensity change of the Canny edge detector, if the amplitude of gradient of one pixel is greater than the width of gradient of the pixels on both sides, the pixel is classified as belonging to an edge.

Texture and color features can be regarded as edge indicators because the complexity of texture and the change of color can be represented by the amplitude of the gradient. Compared with interference frames, normal frames involve more complex textures and more sudden color changes. Therefore, the amplitude of the gradient of pixels in normal frames is greater than that of pixels in interference frames.

\begin{figure}[htbp]
\centerline{\includegraphics[width=21pc]{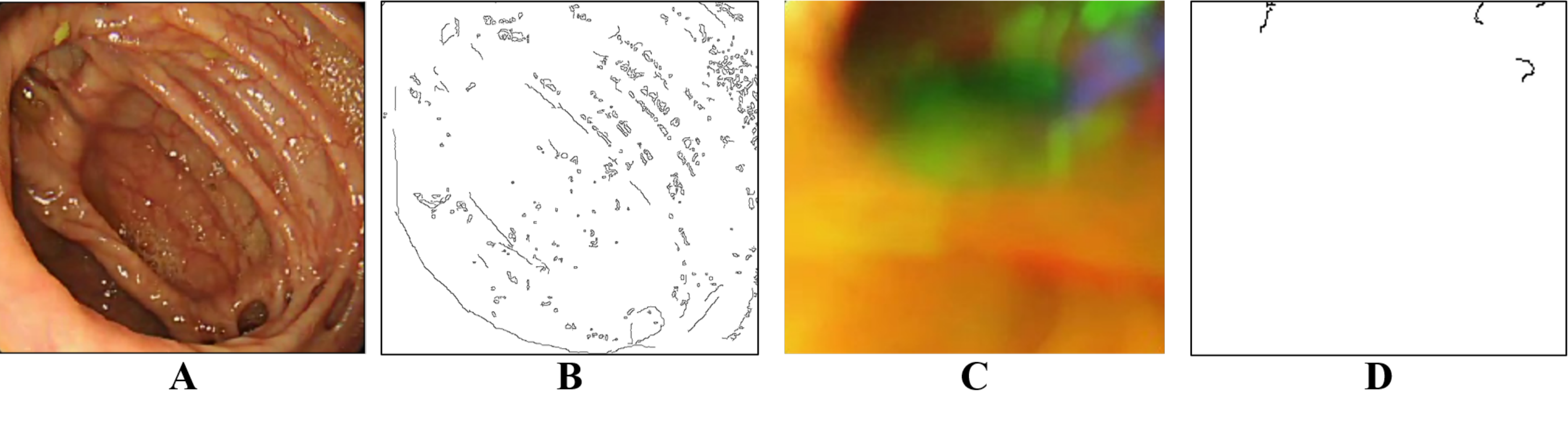}}
\caption{(a) A normal frame in colonoscopy video. (b) The edge feature of normal frame extracted by Canny edge detector. (c) An interference frame in colonoscopy video. (d) The edge feature of interference frame extracted by Canny edge detector.}
\label{fig-method-preprocess}
\end{figure}

As shown in Fig. \ref{fig-method-preprocess}, A is an original normal frame and B depicts the edge features of it, while C is an original interference frame and D depicts the edge features of it, in which the black pixels belong to edges whereas the white pixels do not. It can be observed that the edge pixels in the normal frame are more numerous than the edge pixels in the interference frame. We calculate the number of black edge pixels for each frame that passed through the Canny detector. To experimentally determine the criterion that distinguishes interference frames from normal frames, we perform classification of $200$ images using various choices of threshold and obtained an optimal threshold of $2,000$, which means a frame should be classified as normal if it contains more than $2,000$ black edge pixels. The threshold is then applied to classify outputs of the Canny detector into normal class or interference class and thereby reject interference frames in colonoscopy videos.

\subsection{Landmark Detection Network Based on ResNet-101}
\begin{figure*}[htbp]
\centerline{\includegraphics[width=40pc]{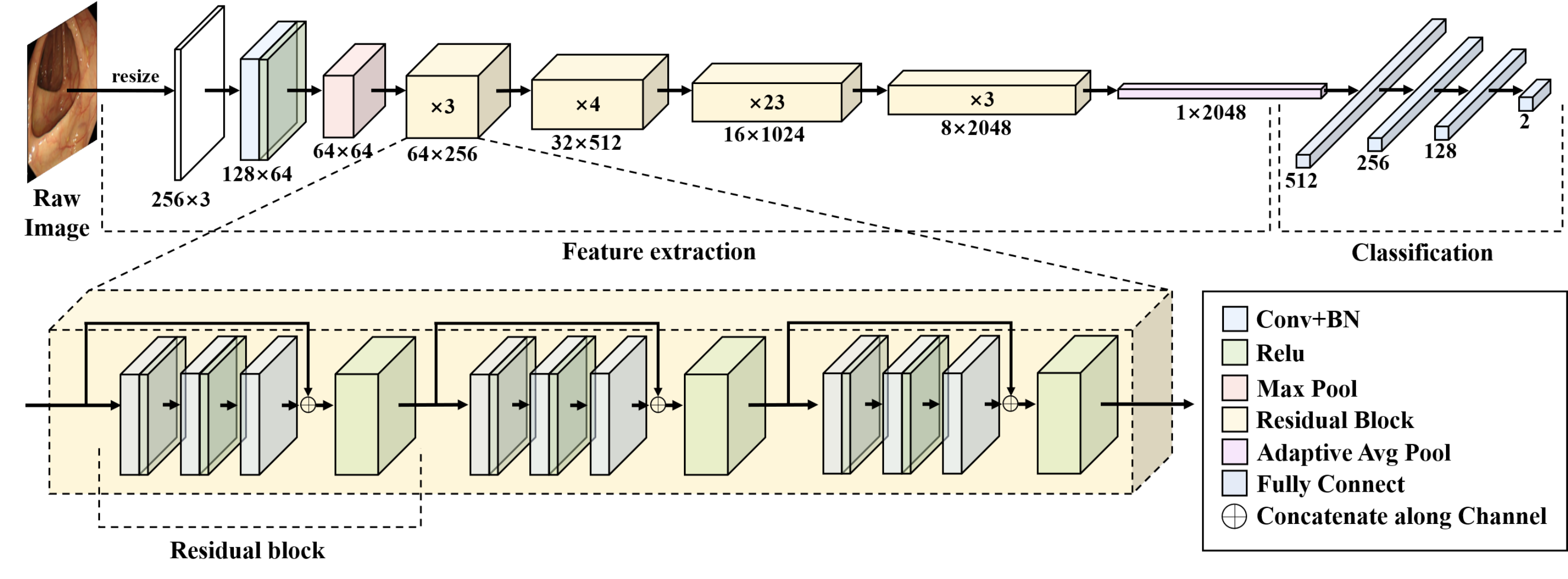}}
\caption{Our proposed landmark detection model based on ResNet-101. The single 1000-way fully connected layer is substituted by four consecutive fully connected layers. A dropout layer is inserted in between each pair of the four fully connected layers, which randomly drops neurons from visible and hidden layers to avoid overfitting.}
\label{fig-resnet}
\end{figure*}
The network inputs are $256\times256$ colonoscopy images, while the network output is a two-dimensional vector indicating whether an input image should belong to the biological anatomical landmark (positive) class or the normal issue (negative) class. The network structure is developed based on the ResNet-101 model. As shown in Fig. \ref{fig-resnet}, the first layer is a $7\times7$ convolutional layer with a stride of $2$. The second layer is a $3\times3$ max pooling layer with a stride of $2$. Next, four different kinds of 3-layer building blocks are stacked 3 times, 4 times, 23 times, and 3 times, respectively. In the original model, the network ends with a global average pooling layer and a 1000-way fully connected layer with LogSoftMax. In the modified model for this task, the single 1000-way fully connected layer is replaced by four consecutive fully connected layers. To prevent overfitting, we use a dropout layer between each pair of the four consecutive fully connected layers to randomly drop neurons from visible and hidden layers. For the three dropout layers, the probabilities of an element to be set as zero are 0.4,0.3,0.3, respectively. 

Due to the considerable depth of the network and the limited volume of training data, it is difficult to train the ResNet-101 model from scratch. Instead of randomly initializing the model weights, we load the weights pre-trained on the ImageNet dataset \cite{5206848} for all layers except the last four fully connected layers. The ImageNet dataset is a public dataset containing over 14 million quality-controlled and human-annotated natural images belonging to 1000 categories.

Output from the last fully connected layer could be activated by LogSoftmax activation, as shown in Eq. \ref{eq-logsoftmax}.
The LogSoftmax formulation can be described as:
\begin{equation}\label{eq-logsoftmax}
    LogSoftmax(x_i)=\log(\frac{\exp(x_i)}{\sum_{j}^N \exp(x_j)})
\end{equation}
where $N$ is the number of classes, which is 2 in our case. $x_i$ represents the specific predicted result of the landmark class or the normal tissue class.


Our loss function is the negative log likelihood loss (NLLLoss), which is useful to train a classification problem. Its input is a log probability vector and a target label. Our reduction is "mean". NLLLoss can be described as:
\begin{equation}
    \begin{aligned}
        l_n &= -x_n y_n\\
        L(x,y) &= \sum_{n=1}^M \frac{l_n}{M}
    \end{aligned}
\end{equation}
where $M$ denotes the batch size, which is $32$ in our case. $x_n$ represents the predicted result of all classes. $x_n y_n$ represents the dot product of vector $x_n$ and vector $y_n$.

Since the colonoscopy images exhibit distinctive features that vastly differ from those of the natural images in the ImageNet dataset, freezing the model weights as pre-trained has the potential to hinder the learning of features from colonoscopy images. Therefore, during the model training stage, the weights of all layers are not frozen. They are updated through the backpropagation of gradients in every epoch.

In total, our proposed anatomical landmark detection model based on ResNet-101 contains $43,713,730$ trainable parameters. It takes 0.0279 seconds on average for the model to detect a single video frame into the biological anatomical landmark class or the normal tissue class. The statistics mentioned above are obtained from experiments conducted on an NVIDIA GeForce RTX 2080 Ti GPU.


\subsection{Post-process: Result Cleaning}
After passing the colonoscopy video frames through the network, we obtain the intermediate detection results indicating whether each frame should be classified as positive or negative. The next step toward the final results is to locate the landmark periods within the whole video period. However, it can be observed that some negative frames are discretely distributed in the presumably continuous landmark periods, while some positive frames are discretely distributed in the non-landmark normal periods. Therefore, it is necessary to post-process the intermediate detection results by identifying the incorrectly predicted frames and reassigning them back to the correct class.

As shown in Fig. \ref{fig-flow}, in the intermediate detection results, within the landmark period, a large proportion of frames are predicted as positive while a small proportion are predicted as negative. In contrast, outside the landmark period, a large proportion of frames are predicted as negative while a small proportion are predicted as positive. Based on the fact that each landmark period should be continuous, the incorrectly predicted frames in intermediate detection results can be identified according to the temporal distribution.

The specific post-processing strategy is as follows. For each frame, we count the number of positive frames among eight neighboring frames, including four on the left and four on the right. If the count is greater than $3$, the frame of interest should be classified as positive; otherwise, it should be classified as negative. For the first and the last four frames, we count the number of positive frames among ten neighboring frames. If the count is greater than $5$, the first and the last four frames should be classified as positive; otherwise, they should be classified as negative. The detailed algorithm is shown in Algorithm \ref{algorithm}.

\begin{algorithm}[!h]
\SetProcArgSty{texttt}
\label{algorithm}
  \caption{Result\_cleaning} 
  \KwIn{Intermediate detection result: $y_{pred}$}  
  \KwOut{Final detection result: $y_{final}$} 

    \SetKwFunction{FMain}{Result\_cleaning}
    \SetKwProg{Fn}{Function}{:}{}
    \Fn{\FMain{$y_{pred}$} }{
        \While{\textbf{True}}{
            $y_{final} \longleftarrow empty\ list$;
            
            \FuncSty{Check\_ends($y_{pred}[:10]$)};
     
            \For{$i = 4 \to$ \FuncSty{len(input\_list)}$-4$} {
                $check\_list \longleftarrow input\_list[i-4:i+4+1]$;
                
                \eIf{\FuncSty{Sum($check\_list[0:4,5:8]>3$)}}
                {${final}.append(1)$;}
                {$y_{final}.append(0)$;}
            }
        
            \FuncSty{Check\_ends($y_{pred}[-10:]$)};
            
            \eIf{\FuncSty{Check\_step($y_{pred}$)} $=1$}
                {\textbf{break};}
                {$y_{pred} \longleftarrow y_{final}$;}
        }
        \textbf{return} $y_{final}$;
    }

\end{algorithm}

\begin{algorithm}[!h]
  \caption{Check\_ends} 
  \KwIn{$input\_list$}
    \SetKwFunction{FMain}{Check\_ends}
    \SetKwProg{Fn}{Function}{:}{}
    \Fn{\FMain{$input\_list$} }{
        \eIf{\FuncSty{Sum($input\_list)>5$)}}
            {$y_{final}.extend([1,1,1,1])$;}
            {$y_{final}.extend([0,0,0,0])$;}
    }
\end{algorithm}

\begin{algorithm}[!h]
  \caption{Check\_step} 
  \KwIn{$input\_list$}
  \KwOut{Number of step: $step$}
    \SetKwFunction{FMain}{Check\_step}
    \SetKwProg{Fn}{Function}{:}{}
    \Fn{\FMain{$input\_list$} }{
        $step \longleftarrow 0;$
        
        \For{$i = 0 \to$ \FuncSty{len(input\_list)}$-1$}{  
            $step+= |input\_list[i]-input\_list[i+1]|$;
        }
        
        \textbf{return} $step$;
    }
\end{algorithm}

\section{Experiment Results}
\label{experimental results}
\subsection{Experimental data and setup}
\label{experimental_data_and_setup}

In our experiment, we introduce a P-N gap between positive and negative periods and sample them with adaptive sampling frequencies. The reason for leaving a P-N gap is that the labels of landmark periods may contain some errors, which blur the boundaries between positive and negative samples. Therefore, when parts of the positive and negative samples are very similar, it is possible for the model to learn from the wrong samples. Another crucial issue is regarding avoiding sample imbalance, which would make our detection model suffer from serious bias. To keep the balance between positive and negative samples, we sample them from the videos with adaptive sampling frequencies to ensure that the ratio of positive and negative samples is approximately 1:1, as shown in Fig. \ref{fig-flow}. We set positive sampling frequency as 10 fps, and set negative sampling frequency depending on the formula:

\begin{equation}
    f_{n} = f_{p}\times\frac{t_{p}}{t_{n}}
\end{equation}

where $f_{n}$ and $f_{p}$ denote negative and positive frequencies, and $t_{n}$ and $t_{p}$ denote negative and positive time periods in the video.

The training set contains 80\% of the total samples, while the test set contains 20\%. We also create a validation set from the training set to compare different network performances and to prevent overfitting. Details are shown in Table \ref{table-data}, in which L1, L2, and L3 represent hepatic flexure, splenic flexure, and sigmoid-descending colon junction, respectively. 

\renewcommand\arraystretch{1.2}
\begin{table}[h]
\centering
\setlength{\abovecaptionskip}{0pt}%
\setlength{\belowcaptionskip}{4pt}%
\caption{Sizes of training, validation and test data}
\begin{tabular}{c|c c c|c c|c} 
\hline
& {\it training} & {\it validation} & {\it total} & {\it positive}& {\it negative} &  {\it test}\\
\hline
L1 & 2536 & 634 & 3170 & 1400 & 1770 & 793\\
L2 & 1686 & 421 & 2107 & 957 & 1150 & 527\\
L3 & 1307 & 327 & 1634 & 742 & 892 & 409 \\
\hline
\end{tabular}
\label{table-data}
\end{table}

\subsection{Performance Metrics}
\subsubsection{Accuracy}
Accuracy is defined as the number of correct predictions divided by the total number of predictions, which can be expressed as:

\begin{equation}
    Accuracy = \frac{TP+TN}{TP+TN+FP+FN}
\end{equation}
where true positive (TP) cases are the biological anatomical landmarks frames that are correctly predicted as landmark while true negative (TN) cases represent the normal frames that are correctly predicted as normal. False positive (FP) cases are the normal frames incorrectly predicted as landmark while false negative (FN) cases represent the biological anatomical landmarks frames which are incorrectly predicted as normal.

\subsubsection{Precision}
Accuracy is not necessarily a dependable indicator of model performance, especially when the class distribution is highly imbalanced, i.e., one class is significantly larger in size than the others. In this case, predicting all samples as the most frequent class would result in a high accuracy rate, yet the seemingly satisfactory number cannot serve as valid proof for the predicting capability of the model. Precision provides a relatively unbiased measure when the class distribution is imbalanced. It is defined as:
\begin{equation}
    Precision = \frac{TP}{TP+FP}
\end{equation}

\subsubsection{Recall}
Recall is another important performance metric defined as the fraction of samples from a specific class that are correctly predicted by the model. Intuitively, precision indicates the ability of the classifier to avoid labeling a negative sample as positive, while recall indicates the ability of the classifier to identify all the positive samples. The formal definition of recall is given by:

\begin{equation}
    Recall = \frac{TP}{TP+FN}
\end{equation}

\subsubsection{F1}
F1 score is a performance metric that balances precision and recall by calculating the harmonic mean between them. F1 score takes its value within the range between 0 and 1, with a greater value indicating better model performance. It evaluates both the preciseness and the robustness of the classifier by considering the number of instances it correctly classifies and the number of instances it misses. Mathematically, F1 score is expressed as:

\begin{equation}
    F1 = \frac{2}{\frac{1}{precision}+\frac{1}{recall}}
\end{equation}

\subsubsection{IoU}
IoU (Intersection over Union) is a metric frequently used to evaluate the extent of overlapping of two regions in applications related to object detection. IoU takes its value within the range between 0 and 1, with a greater value indicating a more considerable extent of overlapping. Two precisely overlapped regions yield an IoU of 1, while two disjointed regions yield an IoU of 0.

\begin{figure}[htbp]
\centerline{\includegraphics[width=21pc]{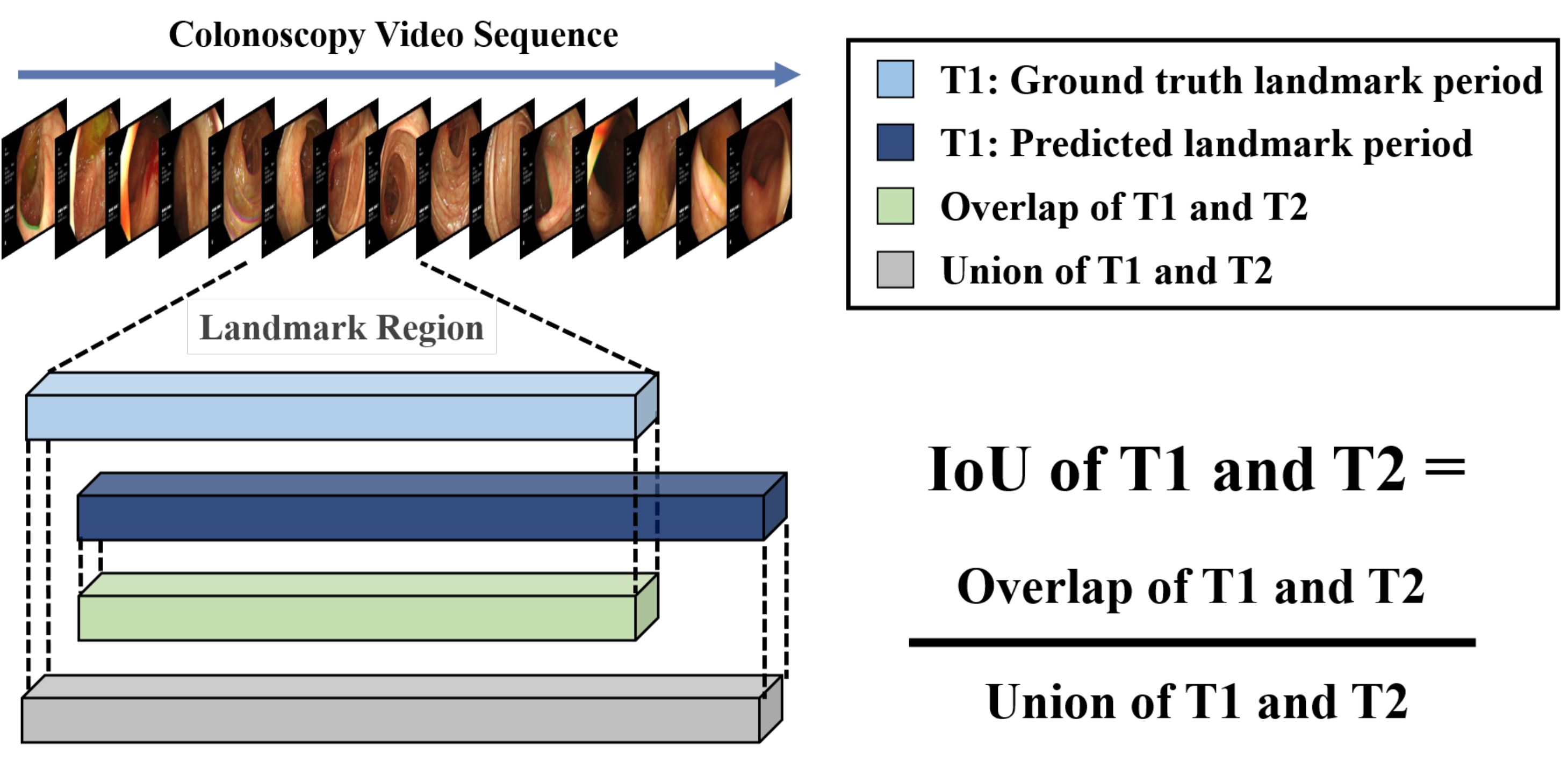}}
\caption{IoU of two time periods}
\label{fig-iou}
\end{figure}

The traditional definition of IoU is to evaluate the similarity between our predicted bounding box and the ground truth bounding box in object detection. Based on the IoU defined in two-dimensional spatial context, we derive the IoU defined in one-dimensional temporal context. As shown in Fig \ref{fig-iou}, the IoU of two time periods, T1 and T2, can be obtained by dividing the length of their overlapping period over the length of their union period.

\subsection{Training and Implementation Details}
We train three separate models to detect three biological anatomical landmarks respectively. The original colonoscopy images have the original resolution of $521\times478$ reduced to $256\times256$ to reach a trade-off between detection accuracy and computational cost. For our model, we use Adam optimizer during training with a learning rate of $0.001$. The batch size is $32$, and the number of epochs is $200$.

The test set here is the video clip between the two landmarks described above as shown in Fig. \ref{fig-flow}, which contains 20\% of the total samples. We input these test samples into the trained detection model and classify them into positive class or negative class. As we mentioned in Section \ref{method}, after obtaining the intermediate detection results, we post-process them to retrieve the final detection results, which also include the overlapping information of predicted landmarks and ground truth. We detect and locate all the three landmarks following the above pipeline.

\subsection{Intermediate Detection Result}

We utilize the above parameters for training the detection model. In the implementation, the test set is as described above, and the various test indicators for the three landmarks are shown in Table \ref{table-class}. 

\renewcommand\arraystretch{1.2}
\begin{table}[htbp]
\centering
\setlength{\abovecaptionskip}{0pt}%
\setlength{\belowcaptionskip}{4pt}%
\caption{Intermediate detection results in terms of multiple test indicators}
\begin{tabular}{c|c c c c} 
\cline{1-5}
& {\it Accuracy} & {\it Precision} & {\it Recall} & {\it F1 score}\\
\hline
L1& 0.9072 & 0.9092 & 0.9077 & 0.9071\\
L2& 0.9085 & 0.9170 & 0.8751 & 0.8916\\
L3& 0.9203 & 0.9234 & 0.9101 & 0.9157\\
\hline
\end{tabular}
\label{table-class}
\end{table}


In terms of intermediate detection accuracy, our proposed model based on ResNet-101 reaches 90.72\%, 90.85\%, and 92.0\% for the three landmarks, respectively. Furthermore, the results of precision, recall, and F1 score illustrate that our proposed model guarantees the balance between precision and recall, which demonstrate that our detection model is both precise and robust.

\subsection{Comparison with other models}
To further evaluate the performance of our proposed model based on ResNet-101, we carry out comparative experiments with four other frequently applied deep learning models, including vgg16, Inception v3, and ResNet-50, and ResNet-101. We implement these models on our dataset and summarize the corresponding results of accuracy, precision, recall, and F1 score in Table \ref{table-compare}.

\renewcommand\arraystretch{1.2}
\begin{table}[!htbp]
\centering
\setlength{\abovecaptionskip}{0pt}%
\setlength{\belowcaptionskip}{4pt}%
\caption{Comparison with other models in intermediate detection results}
\begin{tabular}{c|c|c c c c} 
\cline{1-6}
& & {\it Accuracy} & {\it Precision} & {\it Recall} & {\it F1 score}\\
\hline
\multirow{4}*{L1}& vgg16 & 0.5058 & 0.2529 & 0.5000 & 0.3359\\  
& Inception v3 & 0.7587 & 0.7735 & 0.7573 & 0.7547\\
& ResNet-50 & 0.7703 & 0.7952 & 0.7720 & 0.7661\\

& ResNet-101 & 0.8677 & 0.8904 & 0.8682 & 0.8676\\
& {\bf Ours} & {\bf 0.9072} & {\bf 0.9092} & {\bf 0.9077} & {\bf 0.9071}\\
\hline
\multirow{4}*{L2}& vgg16 & 0.6690 & 0.3345 & 0.5000 & 0.4008\\  
& Inception v3 & 0.7923 & 0.8179 & 0.7050 & 0.7243\\
& ResNet-50 & 0.7606 & 0.7647 & 0.6679 & 0.6810\\
& ResNet-101 & 0.8486 & 0.8287 & 0.8304 & 0.8295\\

& {\bf Ours} & {\bf 0.9085} & {\bf 0.9170} & {\bf 0.8751} & {\bf 0.8916}\\
\hline
\multirow{4}*{L3}& vgg16 & 0.6016 & 0.3008 & 0.5000 & 0.3756\\  
& Inception v3 & 0.8167 & 0.8513 & 0.7784 & 0.7917\\
& ResNet-50 & 0.8406 & 0.8527 & 0.8152 & 0.8259\\
& ResNet-101 & 0.9004 & 0.9010 & 0.8902 & 0.8948\\
& {\bf Ours} & {\bf 0.9203} & {\bf 0.9234} & {\bf 0.9101} & {\bf 0.9157}\\
\hline
\end{tabular}
\label{table-compare}
\end{table}

As shown in Table \ref{table-compare}, in terms of intermediate detection accuracy, our proposed model outperforms the vgg16 model by 40.14\%, 23.95\%, and 31.87\% for the three anatomic landmarks, respectively. Our proposed model also outperforms the Inception v3 model by 14.85\%, 11.62\%, and 10.36\%, while outperforms the ResNet-50 model by 13.69\%, 14.79\%, and 7.97\% for the three anatomic landmarks, respectively. As for ResNet-101, our proposed model outperforms it by 3.95\%, 5.99\%, and 1.99\%, respectively. In addition, the comparison results of precision, recall, and F1 score further demonstrate the superiority of our proposed model over the others in anatomic landmarks recognition. The comparison among vgg16, Inception v3, and the proposed model indicates the importance of residual blocks in network architecture, while the comparison among ResNet-50, ResNet-101, and the proposed model indicates that increasing the network depth could contribute to performance improvement.

\subsection{Final Detection Result}
To better locate the landmark periods within the whole video period, we propose to post-process the intermediate detection results by identifying the incorrectly predicted frames based on their temporal distribution and reassigning them back to the correct class.

\renewcommand\arraystretch{1.2}
\begin{table}[htbp]
\centering
\setlength{\abovecaptionskip}{0pt}%
\setlength{\belowcaptionskip}{4pt}%
\caption{Final detection results in terms of Accuracy and IoU}

\begin{tabular}{c|c c c} 
\cline{1-4}
& {\it Intermediate Acc} & {\it Final Acc} & {\it Final IOU}\\
\hline
L1 & 0.9072 & 0.9974 & 0.8960\\
L2 & 0.9085 & 0.9980 & 0.9039\\
L3 & 0.9203 & 0.9971 & 0.9332\\
\hline
\end{tabular}
\label{table-final}
\end{table}

As shown in Table \ref{table-final}, the final detection accuracy for the three landmarks reach 99.74\%, 99.80\%, and 99.71\%, respectively, with an improvement of 9.02\%, 8.95\%, and 7.68\% compared with the intermediate detection accuracy. The results demonstrate that our result cleaning algorithm is capable of distinguishing the wrongly classified frames based on their distribution and then correcting them. 

Meanwhile, to better evaluate the localization performance of the landmarks, we apply the IoU metric to measure the overlap between our predicted landmark periods and ground truth periods. The IoU for the three landmarks reach 0.90, 0.90, and 0.93, respectively, showing a considerable extent of overlap. The final detection results demonstrate that our proposed system is able to accurately detect and localize biological anatomical landmarks against neighboring normal tissue.

\section{Conclusion and Future Work}
\label{conclusion}

To automatically detect biological anatomical landmarks in colonoscopy videos, we present in this article a novel deep learning-based architecture, which gains accuracy from a finely-designed model based on ResNet-101 and computer vision techniques including Canny edge detector and residual learning.
Comprehensive experimental results demonstrate that our proposed ResNet-101 based model outperforms other deep learning-based models in terms of accuracy, precision, recall rate, and F1 score.
Quantitative results indicate that our proposed architecture is capable of correctly differentiating biological anatomical landmarks from neighboring normal regions with an average accuracy of 99.74\%.

In the future, there exist many promising research directions.
First, we will investigate the possibility of applying other state-of-the-art models, including generative adversarial networks (GAN) \cite{goodfellow2014generative}, recurrent neural networks (RNN) \cite{hochreiter1997long} and their variants \cite{mirza2014conditional}\cite{malhotra2015long} to improve the detection accuracy of biological anatomical landmarks in colonoscopy videos.
Second, we will develop a novel positioning algorithm based on the combination of visual information and magnetic trajectory information to estimate the relative distances between lesion areas and detected biological anatomical landmarks.
Furthermore, another challenging future extension is to establish a 3D space that restores the internal structure of the colon based on 3D reconstruction technologies \cite{izadi2011kinectfusion}, which has the potential to significantly improve the efficiency of diagnosing lesion areas.

\bibliographystyle{ieeetr} 
\end{document}